\documentclass[10pt,twocolumn,letterpaper]{article}

\usepackage{wacv}
\usepackage{times}
\usepackage{epsfig}
\usepackage{graphicx}
\usepackage{amsmath}
\usepackage{amssymb}
\usepackage{booktabs}
\usepackage[accsupp]{axessibility}  

\usepackage{enumitem}
\wacvalgorithmstrack
\wacvfinalcopy

\ifwacvfinal
\usepackage[breaklinks=true,bookmarks=false]{hyperref}
\else
\usepackage[pagebackref=true,breaklinks=true,colorlinks,bookmarks=false]{hyperref}
\fi

\usepackage{multirow}

\begin{document}

\title{PIDS: Joint Point Interaction-Dimension Search for 3D Point Cloud}
\author{
Tunhou Zhang\textsuperscript{\rm 1}, 
Mingyuan Ma\textsuperscript{\rm 1}, 
Feng Yan\textsuperscript{\rm 2}, 
Hai Li\textsuperscript{\rm 1}, 
Yiran Chen\textsuperscript{\rm 1} 
\\
\textsuperscript{\rm 1}ECE Department, Duke University, Durham, NC 27708\\
\textsuperscript{\rm 2}Department of Computer Science, University of Houston, Houston, TX 77204\\
\textsuperscript{\rm 1} \{tunhou.zhang,mingyuan.ma,hai.li,yiran.chen\}@duke.edu,\\
\textsuperscript{\rm 2} fyan5@central.uh.edu
}

\maketitle
\thispagestyle{empty}

\begin{abstract}
The interaction and dimension of points are two important axes in designing point operators to serve hierarchical 3D models. Yet, these two axes are heterogeneous and challenging to fully explore.
Existing works craft point operator under a single axis and reuse the crafted operator in all parts of 3D models.
This overlooks the opportunity to better combine point interactions and dimensions by exploiting varying geometry/density of 3D point clouds.
In this work, we establish PIDS, a novel paradigm to jointly explore point interactions and point dimensions to serve semantic segmentation on point cloud data. 
We establish a large search space to jointly consider versatile point interactions and point dimensions. This supports point operators with various geometry/density considerations.
The enlarged search space with heterogeneous search components calls for a better ranking of candidate models.
To achieve this, we improve the search space exploration by leveraging predictor-based Neural Architecture Search (NAS), and enhance the quality of prediction by assigning unique encoding to heterogeneous search components based on their priors.
We thoroughly evaluate the networks crafted by PIDS on two semantic segmentation benchmarks, showing $\sim1$\% mIOU improvement on SemanticKITTI and S3DIS over state-of-the-art 3D models. 
\end{abstract}

\section{Introduction}
The rise of 3D acquisition technologies and the increasing amount of geometric data collected from 3D sensors drive the booming of 3D point cloud applications, such as object recognition~\cite{wu20153d}, shape segmentation~\cite{yu2004mesh}, and indoor-scene segmentation~\cite{armeni20163d}.  
Deep Neural Networks (DNNs) plays a critical role in processing 3D point cloud data in an end-to-end fashion~\cite{qi2017pointnet}.
The recent development of neural architecture engineering drives better performance~\cite{zhou2020cylinder3d,zhao2021point} and higher efficiency~\cite{hu2020randla} for 3D point cloud models.

A hierarchical 3D model for point cloud is composed of several point operators. Researchers have identified two heterogeneous axes of a point operator: the interaction of points in a point cloud (point interaction), and the size dimension of the point operator (point dimension).
Starting from PointNet~\cite{qi2017pointnet, qi2017pointnet++}, point-based models employ various strategies to design a point interaction to extract features and make dense predictions for semantic segmentation on 3D point cloud.
Later works propose Multi-layer Perceptron~\cite{qi2017pointnet,qi2017pointnet++}, point convolutions~\cite{xu2018spidercnn,atzmon2018point}, graph neural networks~\cite{wang2019dynamic}, attention mechanism~\cite{zhao2021point,guo2021pct,zhang2022patchformer}, kernel point convolutions~\cite{thomas2019kpconv}, voxelization~\cite{choy20194d}, and 2D projections~\cite{cortinhal2020salsanext,goyal2021revisiting} to improve the  performance-efficiency trade-off on 3D point cloud, with manually tuned point dimensions in different hierarchies of a 3D model.
More recent works leverage Neural Architecture Search (NAS) to explore a wider range of 3D model choices, such as point dimensions in voxelized point interactions~\cite{tang2020searching} and the structural wiring of convolution-based point interactions~\cite{nie2021differentiable}.
 
\begin{table}[t]
    \caption{Comparison of different methods in seeking optimal 3D point operators. Methods marked with $*$ leverage Neural Architecture Search with design automation.}
    \vspace{-0.5em}
    \begin{center}
    \scalebox{0.69}{
    \begin{tabular}{|c|c|c|c|c|}
        \hline
         \multirow{2}{*}{\textbf{Method}} & \multirow{2}{*}{\textbf{Interaction?}} & \multirow{2}{*}{\textbf{Dimension?}} & \multirow{2}{*}{\textbf{Reuse?}} & \textbf{S3DIS} \\
         & & & & \textbf{mIOU (\%)} \\
         \hline
         \textbf{KPConv~\cite{thomas2019kpconv}} & \checkmark & &  \checkmark & 70.6  \\
         \textbf{PointTransformer~\cite{zhao2021point}} & \checkmark & & \checkmark  & 73.5\\
         \textbf{PointSeaNet~\cite{nie2021differentiable}}$^{*}$ & \checkmark & &  \checkmark & 71.9 \\
         \textbf{PIDS (Ours)}$^{*}$ & \checkmark &\checkmark  & & 74.4 \\
         \hline
    \end{tabular}
    }
    \end{center}
    \vspace{-2.5em}
    \label{tab:cmp_methods}
\end{table}

Yet, existing approaches on crafting 3D operators have several limitations, due to the varying distribution of 3D point cloud. First, existing works only craft a single type of point interaction and reuse it within all point operators of a 3D model, see Table \ref{tab:cmp_methods}. The reuse of point interactions can limit the performance of the crafted 3D operators, due to the varying geometric/density distribution of points. 
Second, existing approaches optimize point interactions and point dimensions separately and only seek an optimal solution on a single axis. Such an approach misses the opportunity to discover a better combination of point interactions and dimensions, and thus limits the efficiency of the crafted point operators. Overall, the crafted models usually have sub-optimal performance versus efficiency trade-off on processing 3D point clouds.

In this paper, we propose a joint interaction-dimension search to resolve the above limitations. We envision that existing NAS approaches may have several challenges in 3D point cloud. 
First, 
even with full design automation powered by NAS, it is challenging to jointly explore point interactions and point dimensions simultaneously and precisely on
3D point clouds with varying distribution. Thus, existing weight-sharing approaches such as differentiable NAS~\cite{liu2018darts} and one-shot NAS~\cite{cai2019once,yu2020bignas} may be infeasible, as it is difficult to learn shared weights over point operators, given the fast-changing 3D point cloud inputs.
Second, heterogeneous search components add to the complexity of architecture-performance landscape, making it more difficult to accurately map architectures towards their ground-truth performance. The above challenges call for a more scalable and accurate NAS that innovates automatic model design on serving 3D applications.

To this end, we propose PIDS (Joint \textbf{P}oint \textbf{I}nteraction-\textbf{D}imension \textbf{S}earch for 3D Point Cloud), a novel paradigm that leverages NAS~\cite{zoph2018learning,cai2019once,wen2020neural} to joint explore point interactions and point dimensions.
In PIDS, we craft a massive search space containing point interactions with versatile geometry/density in 3D point cloud and positional size-based point dimensions. For point interactions, we backbone the search space on kernel point convolution~\cite{thomas2019kpconv}, and introduce high-order point interactions to fit geometric dispositions and density distributions within varying 3D points. For point dimensions, we incorporate the design motif from Inverted Residual Bottleneck (IRB~\cite{sandler2018mobilenetv2}) to enable more efficient size-based search components (i.e., width, depth, and expansion factor).
We guide PIDS with predictor-based NAS~\cite{wen2020neural}. We train a neural predictor to map architecture-performance pairs in PIDS search space. To improve the quality of search, we innovate the design of a \textbf{D}ense-\textbf{S}parse (DS) predictor to encode unique priors to point interactions and point dimensions. DS predictor uses sparse embedding features to encode categorical choice of point interactions, and uses dense features to represent continuous choice of point dimensions. The dense features and sparse features are then interacted to compute cross-term product, yielding a superior neural architecture representation that improves performance prediction.

The joint interaction-dimension search in PIDS enables to discover efficient 3D models, and the DS predictor of PIDS allows a better ranking of candidate models (i.e., up to $0.03$ Kendall $\tau$ improvement and $2.6\%$ higher mIOU on semantic segmentation over prior state-of-the-art predictors) to improve the quality of search.
As a result, PIDS crafts high-performing and efficient 3D architectures on various semantic segmentation benchmarks, such as SemanticKITTI~\cite{behley2019semantickitti} and S3DIS~\cite{armeni20163d}. On S3DIS, PIDS network outperforms the state-of-the-art hand-crafted architecture by $\sim$ 0.9\% higher mIOU and is 3.6 $\times$ parameter-efficient. On SemanticKITTI, PIDS network achieves 0.6\% higher mIOU, saves $7.2\times$ parameters, and reduces $7.4\times$ MACs over state-of-the-art NAS-crafted architecture.

We highlight the contributions of this paper as follows:
\begin{itemize}[noitemsep,leftmargin=*]
    \item We propose a new paradigm, PIDS, to jointly explore point interactions and dimensions in a vast search space on 3D point cloud. PIDS optimizes both axes of the point operator and seeks 3D models to strike a balance in performance-efficiency trade-off.
    \item We leverage predictor-based NAS to accurately model candidate networks in the PIDS search space. We further enhance the quality of performance prediction by proposing Dense-Sparse predictor that encodes unique priors on heterogeneous search components and computes cross-term product to gather a better architecture representation.
    \item The best model discovered by PIDS achieves the state-of-the-art mIOU on SemanticKITTI and S3DIS over existing models, with higher efficiency.
\end{itemize}
\section{Related Work}
\noindent \textbf{Deep Learning for 3D Point-Cloud.}
Point interactions are often achieved via point-based networks~\cite{xu2018spidercnn,hu2020randla,lang2020samplenet,thomas2019kpconv} that carries feature aggregations towards irregular point set grids to benefit from learnable radial function and spherical harmonics in 3D point clouds.
Recent point-based literature on point interactions improves point sampling efficiency~\cite{hu2020randla,lang2020samplenet}, local feature aggregations~\cite{zhao2021point,lin2020fpconv}, manual architecture fabrications~\cite{hu2020jsenet,lei2020seggcn}, improved training protocols~\cite{goyal2021revisiting}, and/or better feature selection~\cite{yan20222dpass,cheng20212} to seek higher model performance and efficiency.
However, most of these works manually design a building block, while overlooking the varying geometry and density distributions (i.e., heterogeneity) of 3D point clouds.
In addition, these manual efforts may not be able to explore a wider search space of the point interactions and miss the opportunity to find innovative architecture patterns.
The above limitations in the size of search space (i.e., point dimensions) may negatively impact the quality and efficiency of the designed models.

\noindent \textbf{Predictor-based NAS.}
Predictor-based NAS~\cite{wen2020neural,dudziak2020brp} is a popular method that trains a performance predictor on limited samples.
The trained predictor serves as a surrogate model of the ground-truth performance and is utilized to guide architecture search on the full design space.
Existing achievements on predictor-based NAS lies in improving sample efficiency~\cite{dudziak2020brp} and augmenting architecture samples~\cite{liu2021homogeneous} of the performance predictors, yet does not emphasize on improving the neural architecture representations to reduce prediction error.
In the design space for 3D models, existing approaches may be subject to weak prediction performance without the correct neural architecture representations given versatile heterogeneous search components, requiring a deep study on the encoding of priors on different heterogeneous search components. 
\section{PIDS Search Space}
\begin{figure}[t]
    \vspace{-2em}
    \begin{center}
    \includegraphics[width=0.9\linewidth]{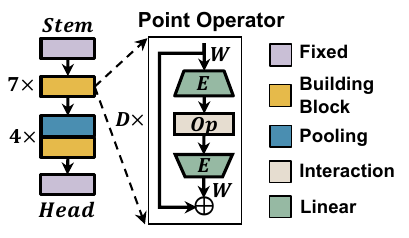}
    \vspace{-0.5em}
    \caption{Overview of PIDS search space. Within each point operator, $Op$ denotes the search component for point interactions. $D$: Depth, $W$: Width and $E$: Expansion Factor denote the search component for point dimensions.}
    \label{fig:designspace}        
    \end{center}
    \vspace{-2em}
\end{figure}

In this section, we formally present PIDS, a new paradigm that jointly searches point interactions and point dimension. Figure \ref{fig:designspace} demonstrates the joint interaction-dimension search space in PIDS. A 3D model for semantic segmentation uses an encoder-decoder structure with 11 searchable stages, with 7 stages in backbone encoder model and 4 stages in segmentation decoder model. 
The searchable stages are sandwiched by fixed stem/head layer to correctly handle inputs/outputs. Each stage is composed of several point operators as basic building blocks. Each point operator has configurable point interactions (e.g., orders of interaction) and point dimensions (e.g., width) in the search space. Intuited by the design motifs from KPConv~\cite{thomas2019kpconv}, PIDS proposes high-order point interactions and seeks the optimal adaptations towards varying geometry and density in different part of the point cloud. Intuited by Inverted Residual Bottleneck (IRB)~\cite{sandler2018mobilenetv2}, PIDS expands the exploration scope for point dimensions (i.e., depth, width, and expansion factor within a point operator) to discover more flexible model choices.

\subsection{High-order Point Interactions}
In a 3D point cloud, a point interaction is defined as a symmetric function over a center point $X$ and its $N$ neighboring points $X_{N}$. 
Given a center point $X\in \mathcal{R}^3$ and its corresponding features with $D$ dimensions: $F \in \mathcal{R}^{D}$, point interaction takes all $N$ neighbor points $X_n\in R^{N \times 3}$ with their corresponding features $F \in \mathbb{R}^{N \times D}$ to compute the output features $F' \in \mathbf{R}^{N \times D}$ via a learnable transformation $f$ parameterized by $\theta$, illustrated as follows:
\begin{equation}
    F' = f(F, X, X_{n}; \theta) .
\end{equation}
We use the intuitions from kernel point convolution~\cite{thomas2019kpconv} to build point interactions. This is because in a kernel point convolution, the neighbor points $X_{n}$ is efficiently achieved by performing radius sampling, taking up only $\sim$30\% of the total computation cost. Thus, optimizing the point interactions and point dimensions shows promising improvement on the overall 3D model.
Based on kernel point convolution, we introduce our first-order and second-order point interactions. A stacking of the first/second point interactions lead to high order point interactions in the 3D model.

\begin{figure}[b]
    \vspace{-4em}
    \begin{center}
    \includegraphics[width=0.9\linewidth]{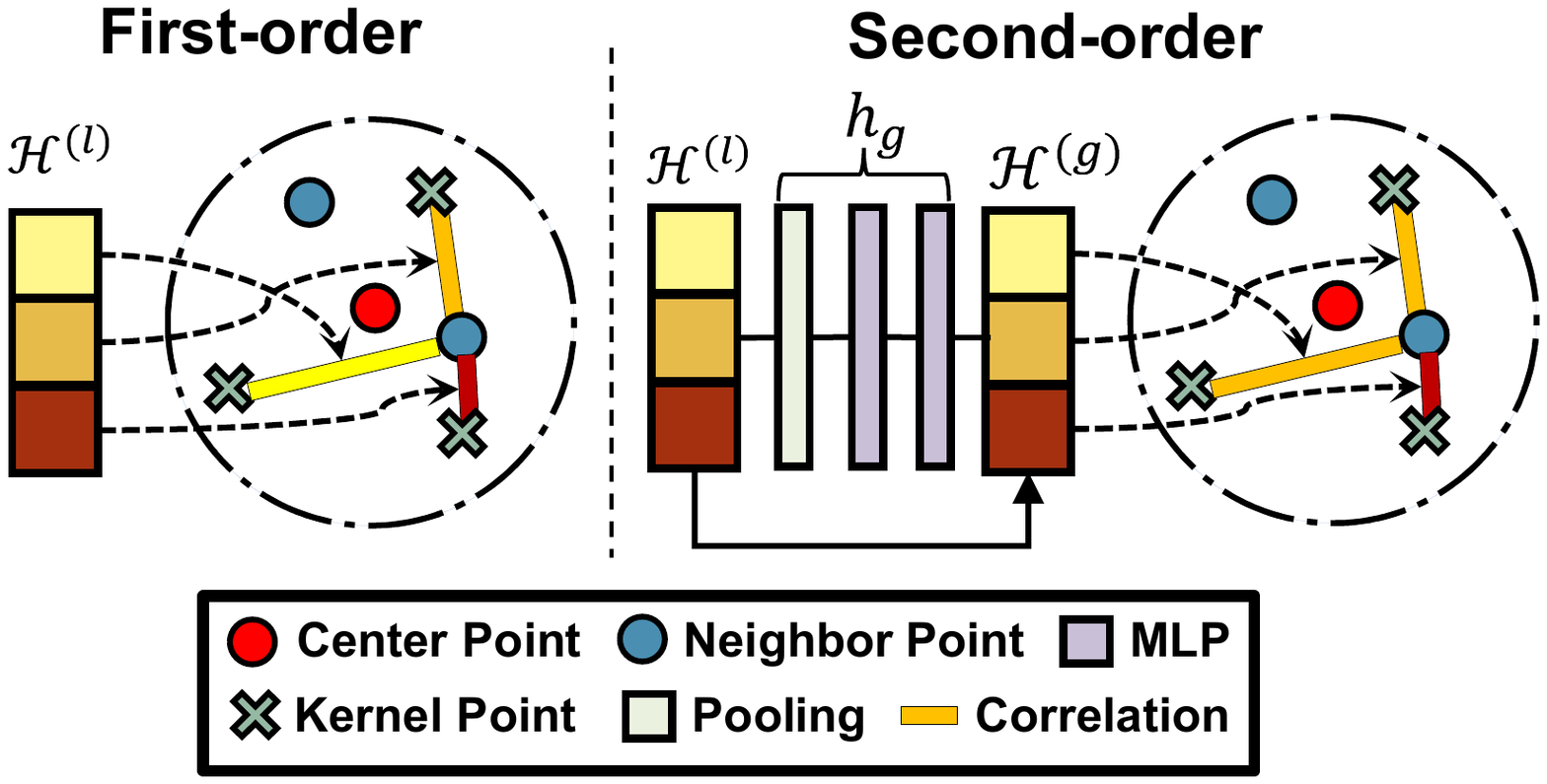}
    \vspace{-3em}
    \caption{First-order and second-order point interaction. The first order point interaction locally models a center point towards a geometric kernel, and the second-order point interaction additionally utilizes global neighbors to capture points with varying density.}
    \label{fig:interact}        
    \end{center}
    \vspace{-1em}
\end{figure}

\noindent \textbf{First-order Point Interaction.}
A first-order point interaction assigns a unique kernel $\mathcal{K}$ with $K$ kernel points that carry weights $W_k \in \mathbb{R}^{K \times D}$ of all feature dimensions.
A first-order point interaction adopts a linear correlation function $h_{l}: \mathbb{R}^{N \times D} \to \mathbb{R}^{K \times D}$ that 
maps the unordered neighbor point features $F$ to a set of features $\bar{\mathbf{F}}=[\bar{F}_{1}, \bar{F}_{2}, ..., \bar{F}_{K}]$ on $K$ kernel points. Specifically, a linear correlation function $h^{(l)}$ measures the contribution of a neighbor point $x_{i} \in X_{n}$ towards a kernel point $\hat{x}_k$ and outputs a linear correlation $\mathcal{H}^{(l)} \in \mathbb{R}^{N \times K}$ as follows:
\begin{equation}
    \mathcal{H}^{(l)}_{i, k} = h^{(l)}(x_i, \hat{x}_k) = \max(0, 1 - \frac{||x_{i} - \hat{x}_{k}||_2}{\delta}),
\end{equation}

Where $\delta$ is a hyperparameter that indicates the influence of a kernel over neighboring points. The first-order interaction aggregates the kernel features $\bar{\textbf{F}}$ via:
\begin{align}
\label{eq:1st_interact}
    f_{1st}(F_{i}) & = \sum_{k}\mathcal{H}^{(l)}\hat{F}_{k}W_{k} \\
\label{eq:1st_interact_2}
                   & = \sum_{x_i \in X_{n}}h^{(l)}(x_{i}, 
                   \hat{x}_k)F_{i}W_{k}, \forall x_{i} \in X_{n},
\end{align}
where $\hat{F}_k \in \mathbb{R^{D}}$ indicates a $D$-dimensional feature mapped to the kernel points, $F_{i}$ denotes an original D-dimensional feature maintained in local neighborhood points. 
Eq. \ref{eq:1st_interact},\ref{eq:1st_interact_2} establish a direct interaction between a center point and its surrounding neighbor points by an explicit form of linear combination, carried by a specific kernel point $W_{k}$ that simultaneously operate on all $D$ features. Different from the original kernel point convolution, first-order point interaction carries element-wise multiplication between weighted features (i.e., $H^{(l)}\hat{F}_{k}$) and kernel weights (i.e., $W_{k}$), without considering the channel-wise features.
This saves $D\times$ parameters and Multiply-Accumulates compared to the original kernel point convolution, yielding higher efficiency in 3D models.

In KPConv, kernel point convolution uses a fixed kernel size (i.e., 15) within all parts of the model, indicating a fixed geometric kernel disposition $\hat{x}_k$ for points in different hierarchical levels of the point cloud. Our first-order interaction extends the scope of kernel dispositions to allow versatile choices in kernel dispositions. Given different kernel dispositions , the first-order point interaction has the ability to capture different geometric properties within the mast 3D point cloud, leading to higher quality of learned representations with fast-changing point cloud geometries. 

\noindent \textbf{Second-order Point Interaction.}
Note that in first-order point interaction, the linear correlation $\mathcal{H}^{(l)}$ that measures the importance of a kernel point towards a neighbor point purely uses the Euclidean distance between them. Such importance measurement overlooks the contribution of other neighbors, thus may not be accurate under point clouds with varying densities. For example, farther kernel points might be more important to neighbor points in a sparse point cloud, where each center point has only a few neighbor points within a sparse distribution of point clouds.

To cover the varying geometric and dense disposition of kernels, we introduce a density-aware correlation $H^{(g)}$ as an alternative to the linear correlation $H_{l}$. The density-aware correlation leverage the global neighborhood information of center points and make adaptations to the varying density of these points. Specifically, the density-aware linear correlation $H^{(g)}$ takes linear correlation $H^{(l)}$ as input, and uses gating function $h_g$ to re-calibrate the kernel-neighbor importance as follows:
\begin{equation} 
    \label{eq:2nd_correlation}
    H^{(g)} = h_{g}(\mathcal{H}^{(l)}) = {\rm \sigma}({\rm MLP}({\rm Pool}(\mathcal{H}^{(l)}))) \odot \mathcal{H}^{(l)},
    \vspace{-0.5em}
\end{equation}
Here, $\sigma$ is the sigmoid function, $\rm MLP$ denotes a 2-layer MLP layer with learnable weights, $Pool$ denotes the global average pooling operator that averages the information in the neighbor dimension of $\mathcal{H}^{(l)}$. 
Based on the density-aware linear correlation $H^{(g)}$, we establish the second-order interaction as follows:
\begin{equation}
\label{eq:2nd_interact}
    f_{2nd}(F_{i}) = \sum_{k} \frac{1}{2} (\mathcal{H}^{(l)} + \mathcal{H}^{(g)})\hat{F}_{k}W_{k}, \forall x_{i} \in X_{n}.
    \vspace{-1em}
\end{equation}

As $H^{(g)}$ is a function of $H^{(l)}$, we employ a summation of $H^{(g)}$ and $H^{(l)}$ for in second-order interaction for easier optimization, following the spirit of residual learning~\cite{he2016deep}. The second-order interaction considers the relative position of neighbor points given a fixed kernel disposition, thus can quickly adapt to 3D point clouds with varying density.
However, second-order point interaction uses slightly more resources due to use of a gating function $h_{g}$, thus should be used sparingly in the design of 3D models for efficiency.

\subsection{Efficient Point Dimensions}
As point interactions only performs point-wise feature extraction, we sandwich high-order point interactions with 2 Fully-connected layer to form a PIDS point operator, see Figure \ref{fig:designspace}.
This enable channel-wise feature interactions in 3D point clouds. Intuited by the design of Inverted Residual Bottleneck (IRB)~\cite{sandler2018mobilenetv2}, we apply an expansion factor on the intermediate point interactions to enrich the representation in low-cost point interaction operations. This IRB design of point operator also provides a wider range of search components for point dimensions.

Next, we validate the effectiveness of the IRB design that combines point interactions and point dimensions in the PIDS point operator. We follow the layer organization of MobileNet-V2~\cite{sandler2018mobilenetv2} to construct a hand-crafted 3D model. We evaluate the crafted 3D model with first-order/second-order point interaction, and compare its performance with KPConv in Table \ref{tab:nkattn_results}.
Even without search, the hand-crafted 3D model with the first-order interaction achieves achieves 0.4\%\ higher mIOU with 12$\times$ smaller size on SemanticKITTI compared to KPConv. The hand-crafted 3D model with second-order interaction pushes the mIOU improvement $\sim 0.5\%$ higher, with only marginal increase in parameter count.
The above empirical evaluations suggest two key insights in designing 3D models: 1) relaxing the choice of kernel dispositions in first-order point interaction provides more opportunities to adapt to geometric distribution of 3D points, and 2) carrying second-order point interaction that leverages global neighborhood information to measure kernel importance makes better adaptations to the varying density of points.

\begin{table}[t]
    \caption{Evaluation of IRB Design on SemanticKITTI.}
    \begin{center}
    \scalebox{0.88}
    {
    \begin{tabular}{|c|c|c|c|c|}
    \hline
    \multirow{2}{*}{\textbf{Architecture}} & \multirow{1}{*}{\textbf{Params}} & \textbf{MACs}  & \multirow{1}{*}{\textbf{SemanticKITTI}} \\
    & \textbf{(M)} & \textbf{(G)} & \multirow{1}{*}{\textbf{mIOU (\%)}} \\
    \hline
    KPConv (Our impl.) & 14.8 & 60.9 & 59.2 \\
    PIDS (first-order) & 0.97 & 4.6 & 59.6 \\
    PIDS (second-order) & 0.98 & 4.7 & 60.1 \\
    \hline
    \end{tabular}
    }
    \vspace{-2em}
    \label{tab:nkattn_results}
    \end{center}
\end{table}

\subsection{Search Components}
In PIDS search space, we jointly search for the best combination of point interactions and point dimensions. We illustrate the two search components as follows:
\vspace{-0.5em}
\begin{itemize}[noitemsep,leftmargin=*]
    \item \textbf{Point Interactions.} We search for the type of interactions (i.e., first-order/second-order) and the choice of geometric kernel dispositions (i.e., 5-point Tetrahedron, 7-point Octahedron, and 13-point Icosahedron dispositions) contained within the point interactions. The choice within point interactions are categorical discrete options. 
    
    \item \textbf{Point Dimensions.} We search for the width (i.e., number of features in each point operator), depth (i.e., the number of stacked of point operators) and expansion factor (i.e., ratio of point interactions to FC operators) of a point operator. The choice within point dimensions are continuous floating-point numbers.
\end{itemize}
\vspace{-0.5em}

Unlike NAS for image-based models that emphasize the search of size-based homogeneous components, the choice of point interactions and the choice of point dimensions are heterogeneous.
In addition, with a total of 11 stages, the overall search space size contains up to  1.8$\times$10$^{19}$ possible architectures, making the architecture exploration under moderate search cost more challenging.
This calls for a precise NAS method to precisely explore the joint point interaction-dimension search space.
\section{Joint Point Interaction-Dimension Search}
We employ predictor-based NAS to jointly explore the best combination of point interactions and point dimensions in 3D models for point cloud.
Specifically, we sample a number of architectures from the joint search space, and use their architecture-performance pairs to train a performance predictor.
Unfortunately, the large cardinality of the joint interaction-dimension search space raise challenges in accurately modeling the search space. In this section, we analyze the priors on search components and propose Dense-Sparse (DS) predictor, a novel surrogate model that encode unique priors to different heterogeneous search features to improve the quality of performance prediction. As a result, DS predictor improves ranking of 3D models within the joint interaction-dimension search space.

\begin{figure}[b]
    \begin{center}
    \vspace{-2em}
    \includegraphics[width=0.9\linewidth]{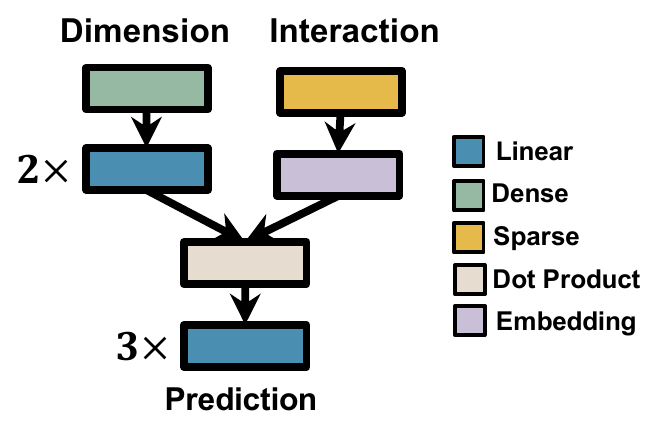}
    \vspace{-1em}
    \caption{Dense-Sparse predictor computes a cross-term product on heterogeneous search components to improve the prediction quality of joint interaction-dimension search.}
    \vspace{-1.25em}
    \label{fig:widedeeppredictor}
    \end{center}
\end{figure}

\subsection{Dense-Sparse Predictor}
We start with analyzing the priors on different search components in PIDS interaction-dimension search space. 
We note that point interactions settings are sparsely distributed within the search space. For example, the choice of kernel dispositions in first/second point interaction follows discrete geometry shapes with no explicit correlation that can be achieved via linear interpolation.
Thus, the choice of point interactions possesses categorical priors which can be implicitly learned by embeddings.
Alternatively, the choice of point dimensions encodes a linear/quasi-linear relationship towards the model performance, as is reflected in the compound scaling scheme in EfficientNet~\cite{tan2019efficientnet}.
Thus, the choice of point interactions contains continuous priors that can be modelled in deep representations.
As a result, using a vanilla predictor to model these multi-modal features with different priors may lead to sub-optimal results.

Motivated by Wide \& Deep learning~\cite{cheng2016wide} that combines memorization and generalization for multi-modality inputs, we propose a novel Dense-Sparse predictor to encode unique priors on point interactions/dimensions. 
This improves the neural architecture representation of candidates in PIDS search space and thus, improves the quality of prediction and ranking of candidate models.
We craft dense/sparse neural architecture representations for point interactions/dimensions, and leverage a dot product to compute the cross-term relationship between point interactions and point dimensions.
Specifically, given a dense representation $X_{d} \in R^{B \times dim}$ and a sparse representation $X_{s} \in R^{B \times N \times dim}$, a dot product carries cross-term feature communication by carrying the following transformation:
\begin{align}
    Z = [vec^{-1}_{B, 1, dim}(X_{d}); X_{s}], \\
    DP(Z) = Triu(ZZ^{T}),
\end{align}

Where $DP$ denotes the dot product operation, and $Triu$ denotes the upper triangular matrix. Figure \ref{fig:widedeeppredictor} demonstrates the architecture DS predictor.
DS predictor adopts a separate MLP to learn deep representations on the continuous dense representations (i.e., point dimensions), and adopts an embedding table to learn the categorical choice of sparse representations (i.e., point interactions).
Thanks to the enhanced encoded priors and their cross-term relationship, DS predictor yields a more accurate modeling of the PIDS search space. Next, we elaborate the training and evaluation of our DS Predictor. 

\begin{table*}[t]
    \begin{center}
    \caption{Dense-Sparse Predictor achieves best prediction quality over SemanticKITTI.}
    \scalebox{1.0}{
    \begin{tabular}{|c|c|c|c|c|c|}
         \hline
         \textbf{Predictor} & \textbf{Rank Loss?} & \textbf{Priors}  & \textbf{Pre-training?} & \textbf{MSE} ($10^{-2}$) & \textbf{Kendall $\tau$}\\
         \hline
         \textbf{Random Forest} & & Continuous & & 3.75$\pm$0.30 & 0.240$\pm$0.044 \\
         \textbf{GPR} & & Continuous & & 4.29$\pm$0.41 & 0.144$\pm$0.042\\
         \textbf{XGBoost} &  & Continuous & & 4.24$\pm$0.40 & 0.210$\pm$0.037 \\
         \textbf{NGBoost} & & Continuous  & & 3.90$\pm$0.33 & 0.255$\pm$0.047\\
         \textbf{LGBoost} & & Continuous & & 4.03$\pm$0.36 & 0.236$\pm$0.033\\
         \hline
         \textbf{Dense} & \checkmark & Continuous & \checkmark & 3.07$\pm$0.33 & 0.379$\pm$0.050 \\ 
         \textbf{Sparse} & \checkmark & Discrete & \checkmark & 2.87$\pm$0.21 & 0.400$\pm$0.047 \\
         \textbf{Dense-Sparse}  & \checkmark & Continuous + Discrete & \checkmark & 2.80$\pm$0.23 & 0.408$\pm$0.044 \\
         \hline
    \end{tabular}}
    \vspace{-2.5em}
    \label{tab:predictor_cmp}
    \end{center}
\end{table*}

\noindent \textbf{Architecture Sampling.}
We randomly sample $\sim$1K architectures on SemanticKITTI.
Following common settings in NAS, we split the training dataset of SemanticKITTI into mini-train and mini-val. 
We train the sampled architectures on mini-train until convergence, and evaluate these architectures on mini-val to obtain architecture performance.

\noindent \textbf{Predictor Training.}
Mean-Squared Error (MSE) lossis the training objective of the DS predictor on 1K sampled architectures.
To ensure a consistent comparison of predictor performance over multiple benchmarks, we carry optimization on predictors with normalized performance.
Since Dense/Sparse/Dense-Sparse predictor can perform gradient-based optimization, we incorporate predictor pre-training on Multiply-Accumulates (MACs) prediction~\cite{dai2021fbnetv3}, and use pair-wise margin rank loss~\cite{dudziak2020brp} to improve ranking.

We compare DS predictor with top-performing predictor designs collected from NASBench-301~\cite{siems2020bench}, see Table \ref{tab:predictor_cmp}.
Since sparse embedding is not feasible in Bayesian models (e.g., GPR) and tree-based methods (e.g., Random Forest), all categorical features are treated as continuous features during the training process.
By enhancing search priors and computing cross-term feature relationship via a dot-product, DS predictor achieves better prediction quality and ranking on PIDS search space: DS predictor achieves up to $0.172$ higher Kendall $\tau$ compared to prior arts, and $0.008$ higher Kendall $\tau$ over Dense predictor that converts all search components to continuous features. 

\subsection{Evolutionary Search on DS Predictor}
With a trained DS predictor, we follow regularized evolution~\cite{real2019regularized} to effectively probe the PIDS search space to carry a joint search on point interactions and point dimensions.
Given a candidate architecture $A$ with performance prediction $\hat{P}$ and MACs $M$, we use $S(A)=\hat{P} - \beta \times \log{M}$ as our search objective, and empirically set $\beta$ to 0.5 to trade-off performance and resource.
We introduce a single mutation of an architecture genotype with the following sequence of actions in one of the stages:
\vspace{-0.5em}
\begin{itemize}[noitemsep,leftmargin=*]
    \item Change the kernel disposition in point operators.
    \item Change the order of interaction in point operators.
    \item Change the width/expansion factor in point operators.
    \item Change the depth of stacked point operators.
\end{itemize}
\vspace{-0.5em}
We use a population of 200 and a sample size of 150 to carry regularized evolution with 360 rounds over PIDS search space to craft NAS models. 

\section{Experiments}
In this section, we evaluate PIDS over two semantic segmentation benchmarks, SemanticKITTI~\cite{behley2019semantickitti} and S3DIS~\cite{armeni20163d}.
Specifically, we carry an end-to-end search on SemanticKITTI to identify the best 3D architecture in PIDS design space, and transfer it to S3DIS.



\begin{table*}[t]
\begin{center}
    \vspace{-1.5em}
    \caption{mIOU on sequence 08 \textbf{(validation split)} of SemanticKITTI. Latency is measured with NVIDIA TITAN X Pascal on a single scene containing $\sim$60K points. Here, \textcolor{red}{red}/\textcolor{blue}{blue} numbers denote computation/processing time. $^{+}$: Results from ~\cite{zhou2021panoptic}. $^{*}$: Results from ~\cite{tang2020searching}.}
    \scalebox{1.0}{
    \begin{tabular}{|c|c|c|c|c|c|}
        \hline
         \textbf{Architecture} & \textbf{Method} & \textbf{Params} & \textbf{MACs} & \textbf{Latency} & \textbf{mIOU} \\
         & & \textbf{(M)} & \textbf{(G)} & \textbf{(ms)} & \textbf{(\%)} \\
         \hline
         RandLANet~\cite{hu2020randla} & Point-based & 1.24 & - & 103 & 57.1 \\
         KPConv-rigid (our impl.) & Point-based & 14.8 & 60.9 & 221 (\textcolor{red}{164} + \textcolor{blue}{57}) & 59.2 \\
         PolarNet~\cite{zhang2020polarnet} & Projection-based & 13.6 & 135.0* & \textbf{62}$^{*}$ & 58.2$^{+}$ \\
         SalsaNext~\cite{cortinhal2020salsanext} & Projection-based & 6.7 & 62.8* & 71$^{*}$ & 59.0$^{+}$ \\
         MinkowskiNet~\cite{choy20194d} & Voxel-based & 5.5 & 28.5$^{*}$ & 294$^{*}$ & 58.9 \\
         PIDS (second-order) & Point-based & \textbf{0.97} & \textbf{4.7} & 160 (\textcolor{red}{103} + \textcolor{blue}{57}) & \textbf{60.1} \\
         \hline
         \multirow{2}{*}{SPVNAS~\cite{tang2020searching}} & \multirow{2}{*}{Voxel-based} & 3.3 & 20.0 & \textbf{158} & 61.5 \\
         & & 7.0 & 34.7 & \textbf{175} & 63.5 \\
         PIDS (NAS) & \multirow{2}{*}{Point-based} & \textbf{0.57} & \textbf{4.4} & 169 (\textcolor{red}{112} + \textcolor{blue}{57}) & 62.4\\
         PIDS (NAS, 2$\times$) & & 1.36 & 11.0 & 206 (\textcolor{red}{149} + \textcolor{blue}{57}) & \textbf{64.1}\\
         \hline
    \end{tabular}}
    \label{tab:result_semantickitti}
    \end{center}
    \vspace{-1.5em}
\end{table*}

\subsection{Hyperparameter Settings}
We describe the detailed hyperparameter settings on SemanticKITTI and S3DIS as follows.

\noindent \textbf{SemanticKITTI.} A single training batch contains 10 sub-sampled point clouds. On SemanticKITTI, we measure Multiply-Accumulates (MACs) over a scene with an average of ~$\sim$12.3K points, which gives a similar MAC count for KPConv architecture as is shown in ~\cite{tang2020searching}. 
We use a downsample rate of $0.06$\textit{m}, and train our best model for 250 epochs with an initial learning rate of 0.04 and cosine learning rate schedule~\cite{loshchilov2016sgdr}. 
We adopt an L2 weight decay of 3e-4 and default data augmentation~\cite{thomas2019kpconv}.

\noindent \textbf{S3DIS.} A single training batch contains 8 sub-sampled point clouds. We use a downsample rate of $0.04$\textit{m} following the original KPConv paper~\cite{thomas2019kpconv}. 
Specifically, we train our best model for 250 epochs with an initial learning rate of 0.04 and cosine learning rate schedule~\cite{loshchilov2016sgdr}.
We adopt a L2 weight decay of 3e-4 and default data augmentation~\cite{thomas2019kpconv}.

\subsection{Evaluation on SemanticKITTI}
We evaluate the top-performing models discovered by PIDS. 
We apply width scaling~\cite{sandler2018mobilenetv2} on the NAS-crafted models and attach a $m\times$ suffix to denote the application of a $m\times$ width multiplier.
This ensures a fair comparison with existing state-of-the-art models of similar sizes.
We compare the performance and efficiency metrics such as parameter count and MACs.

\noindent \textbf{Performance Evaluation.} 
Table \ref{tab:result_semantickitti} shows the mIOU comparison on sequence 08 of the SemanticKITTI dataset. 
Our hand-crafted PIDS model outperforms point-based, projection-based, and voxel-based methods by at least 0.9\%, 1.1\%, and 1.2\% mIOU respectively, with significant parameter and MAC reduction.
The NAS-crafted PIDS model achieves $\sim$ 1\% higher mIOU than the state-of-the-art SPVNAS while saving 5.8$\times$ parameters and 4.5$\times$ MACs. Applying a 2$\times$ width multiplier on NAS-crafted model further boosts the performance, demonstrating a good scalability of discovered 3D models.

\begin{table}[t]
    \vspace{-1em}
    \caption{Operation-level latency breakdown of our searched PIDS model. We list the critical operations here and categorize less important operation as 'Other'. In column \textbf{Type}, \textbf{C/M} denotes computation-bounded/memory-bounded operation.}
    \vspace{-2em}
    \begin{center}
    \scalebox{0.9}{
    \begin{tabular}{|c|c|c|c|}
        \hline
        \textbf{Operation} & \textbf{Type} & \textbf{Latency (ms)} & \textbf{Latency (\%)} \\
        \hline
        Preprocessing (CPU) & \textbf{-} & 57.015 & 33.73 \\
        $\mathtt{aten::sub}$ & \textbf{C} & 23.396 & 13.84 \\
        $\mathtt{aten::bmm}$ & \textbf{C} & 20.236 & 11.97 \\
        $\mathtt{aten::gather}$ & \textbf{M} & 18.115 & 10.72 \\
        $\mathtt{aten::mul}$ & \textbf{C} & 12.487 & 7.39 \\
        $\mathtt{aten::sum}$ & \textbf{C} & 9.887 & 5.85 \\
        $\mathtt{aten::addmm}$ & \textbf{C} & 7.292 & 4.31 \\
        $\mathtt{aten::threshold\_}$ & \textbf{C} & 4.444 & 2.63 \\
        $\mathtt{aten::copy\_}$ & \textbf{M} & 2.912 & 1.72 \\
        $\mathtt{aten::sqrt}$ & \textbf{C} & 2.256 & 1.33 \\
        Other & \textbf{-} & 11.00 & 6.51 \\
        \hline
    \end{tabular}}
    \vspace{-3em}
    \end{center}
    \label{tab:nvprofile_skitti}
\end{table}

\noindent \textbf{Latency Analysis of Best PIDS Model.}
In Table \ref{tab:result_semantickitti}, we notice that our NAS-crafted PIDS model is 4.2 $\times$ faster than original KPConv model despite of its higher mIOU. 
However, we also observe that the latency reduction is less significant compared to MAC reduction of PIDS models, especially compared to voxel-based methods.
This is because voxel-based methods (e.g., SPVNAS~\cite{tang2020searching}) benefit from 1) reduced overhead in point-based neighboring mechanisms 2) mature software-hardware co-design that enables inference with high throughput. 

To dive deep into this issue, we profile the latency breakdown of the NAS-crafted PIDS and demonstrate the profiling result in Table \ref{tab:nvprofile_skitti}.
Here, the preprocessing operations (including radius sampling, grid sub-sampling etc.) consumes 33.7\% of the inference cost, neighbor gathering contributes to the memory-bounded operation (i.e., $\mathtt{aten::gather}$), which takes 10\% of the total inference latency. Computation-bounded parts in KPConv such as depthwise convolution (i.e., $\mathtt{aten::mul}$ and $\mathtt{aten::sum}$), MLPs (i.e., $\mathtt{aten::bmm}$) and point-wise local neighborhood movement (i.e., $\mathtt{aten::sub}$) contribute to $\sim$50\% of the total inference cost.

Besides the optimization of architecture, we also envision some potential improvements that harness the hardware-software co-design to improve the efficiency of PIDS blocks, such as yet not limited to:
\vspace{-0.5em}
\begin{itemize}[noitemsep,leftmargin=*]
    \item Enable efficient radius sampling implementation to allow parallelization between different CPU cores.
    \item Utilize a fused version of convolution (*i.e., $\mathtt{aten::bmm}$ and $\mathtt{aten::sum}$) that aggregates the feature outputs on each kernel point efficiently.
    \item Incorporate the choice of radius for each point operator into the design space and harness such options to discover more efficient point-based 3D models.
    \vspace{-1em}
\end{itemize}

\begin{table}[t]
\vspace{-1em}
    \caption{6-fold cross-validation results on S3DIS.}
    \vspace{-1em}
    \begin{center}
    \scalebox{0.93}{
    \begin{tabular}{|c|c|c|c|c|c|}
        \hline
        \multirow{2}{*}{\textbf{Architecture}} & \textbf{mIOU} & \textbf{mAcc} & \textbf{OA} & \textbf{Params}  \\
        & \textbf{(\%)} & \textbf{(\%)}& \textbf{(\%)} & \textbf{(M)} \\
        \hline
        InterpCNN~\cite{mao2019interpolated} & 66.7 & - & 88.7 & -\\
        RandLANet~\cite{hu2020randla} & 70.0 & 82.0 & 88.0 & 1.24 \\
        KPConv~\cite{thomas2019kpconv} & 70.6 & 79.1 & - & 14.8  \\
        RPNet~\cite{Ran_2021_ICCV} & 70.8 & - & -  & -  \\
        SCF-Net~\cite{Fan_2021_CVPR} & 71.6 & 82.7 & 88.4 & - \\
        BAAF-Net~\cite{Qiu_2021_CVPR} & 72.2 & \textbf{83.1} & 88.9 & \textbf{1.23} \\
        PointTransformer~\cite{zhao2021point} & 73.5 & 81.9 & 90.2 & 4.9 \\
        PointSeaNet~\cite{nie2021differentiable} & 71.9 & - & 90.3 & 6.75 \\
        \hline
        PIDS (NAS, 2$\times$) & \textbf{74.4} & 82.1 & \textbf{90.3} & 1.35 \\
        \hline
    \end{tabular}
    }
    \end{center}
    \vspace{-2em}
    \label{tab:s3dis_6foldcv}
\end{table}

\subsection{Architecture Transferability on S3DIS}

We further verify the transferability of NAS-crafted PIDS models on S3DIS to evaluate its performance on indoor scene segmentation. Following the established protocols in 3D point cloud segmentation, we report both 6-fold cross-validation mIOU across all Area 1$\sim$Area 6 splits, see Table \ref{tab:s3dis_6foldcv}.
By convention, we also report the validation mIOU and mIOU per class on Area 5 to compare with existing methods on 3D point cloud, see Table \ref{tab:S3DIS_mIoU_A5}.

Though our NAS-crafted PIDS model is not optimized on S3DIS, it shows a significant margin on the S3DIS 6-fold cross-validation benchmark compared with existing approaches such as \cite{Qiu_2021_CVPR,zhao2021point} and achieves 74.5\% mIOU, a new state-of-the-art result using only 1.35M parameters (i.e., 3.6$\times$ fewer than the runner-up architecture, Point-Transformer~\cite{zhao2021point}).
On Area 5 of the S3DIS dataset, we observe that: (1) NAS-crafted PIDS achieves 1.8\% higher mIOU than original KPConv with 11$\times$ parameter efficiency. (2) NAS-crafted PIDS achieves competitive performance over different classes on S3DIS indoor semantic segmentation and demonstrates at least $\sim$1\% higher mIOU over 9 out of 13 classes compared to existing approaches.

\begin{table*}[t]
    \caption{mIoU per class on S3DIS Area-5.}
    \centering
    \scalebox{0.9}{
    \begin{tabular}{|c|c|c|c|c|c|c|c|c|c|c|c|c|c|c|c|c|c|c|}
    \hline
        \textbf{Methods} & \textbf{mIoU} & ceil. & floor & wall & beam & col. & wind. & door & chair & table & book. & sofa & board & clut. \\
        \hline
         Pointnet~\cite{qi2017pointnet}      & 41.1 & 88.8 & 97.3 & 69.8 & 0.1 & 3.9  & 46.3 & 10.8 & 52.6 & 58.9 & 40.3 & 5.9  & 26.4 & 33.2  \\
         Eff 3D Conv~\cite{zhang2018efficient}   & 51.8 & 79.8 & 93.9 & 69.0 & \textbf{0.2} & 28.3 & 38.5 & 48.3 & 71.1 & 73.6 & 48.7 & 59.2 & 29.3 & 33.1  \\
         RNN Fusion~\cite{ye20183d} & 57.3 & 92.3 & 98.2 & 79.4 & 0.0 & 17.6 & 22.8 & 62.1 & 74.4 & 80.6 & 31.7 & 66.7 & 62.1 & 56.7  \\
         KPConv~\cite{thomas2019kpconv} & 65.4 & 92.6 & 97.3 & 81.4 & 0.0 & 16.5 & \textbf{54.5} & 69.5 & 90.1 & 80.2 & \textbf{74.6} & \textbf{66.4} & 63.7 & 58.1 \\
         \hline
         PIDS(NAS, 2$\times$)  & \textbf{67.2} & \textbf{93.6} & \textbf{98.3} & \textbf{81.6} & 0.0 & \textbf{32.2} & 51.5 & \textbf{73.2} & \textbf{90.7} & \textbf{82.5} & 73.3 & 64.7 & \textbf{71.6} & \textbf{60.0}  \\
         \hline 
    \end{tabular}
    }
    \label{tab:S3DIS_mIoU_A5}
    \vspace{-1.5em}
\end{table*}
\section{Ablation Studies}
In this section, we first visualize how second-order point interaction re-calibrates the contribution of each neighbor point to suit heterogeneous point clouds with varying density, reflected by the learned coefficients over the geometry disposition of kernels.
Then, we discuss the relationship between MSE reduction of neural predictors, and the quality improvement of discovered NAS models.
\subsection{Second-order Point Interactions: Visualization}
\begin{figure}[b]
  \begin{center}
    \vspace{-4em}
    \includegraphics[width=0.45\textwidth]{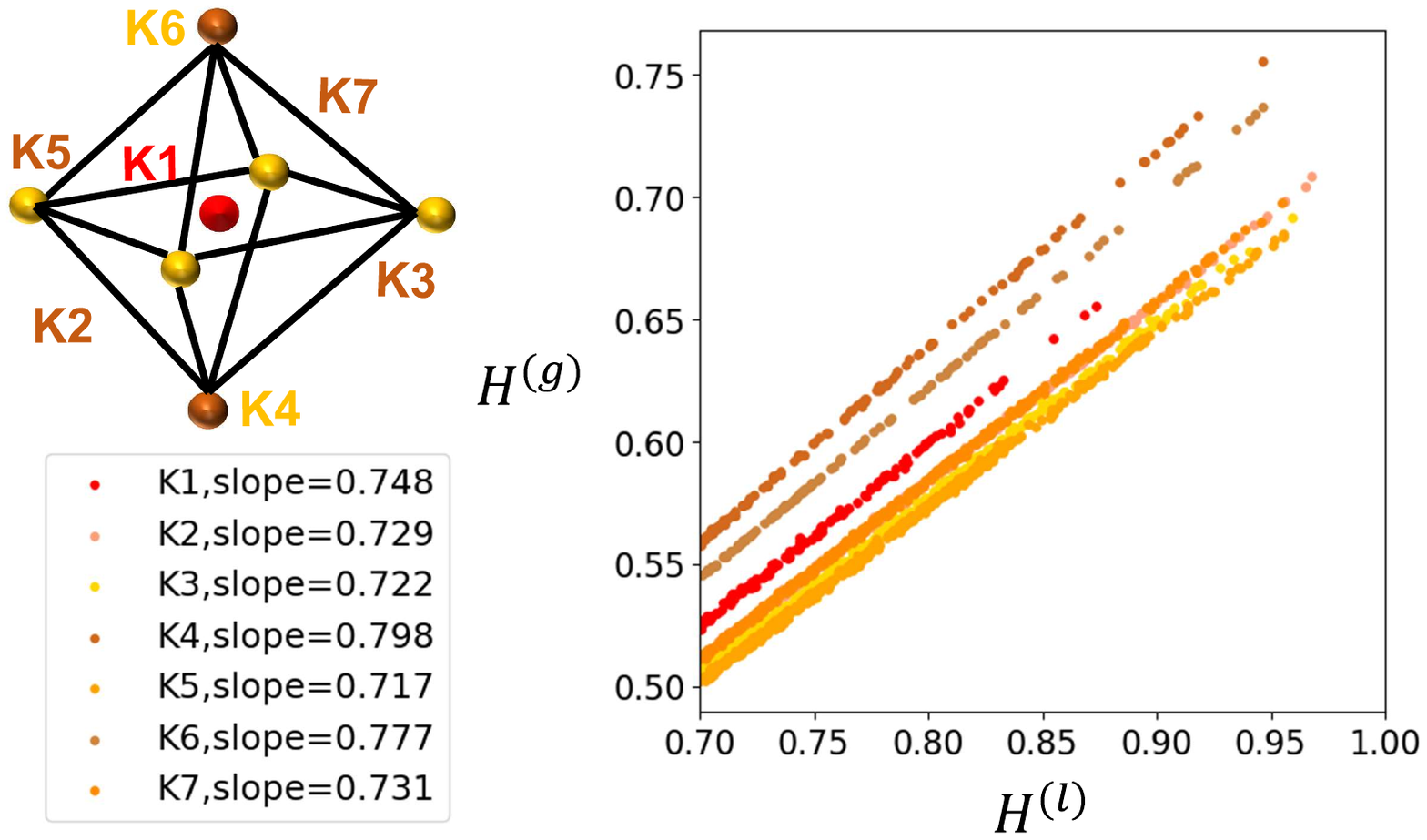}
    \vspace{-2em}
    \caption{Visualization of the second-order point interaction on 7-point Octahedron kernel dispositions.}
  \vspace{-1em}
  \label{fig:attn_visualization}
  \end{center}
\end{figure}

To verify the source of improvement from point interactions, we plot the dynamics of learned point interactions on a 7-point Octahedron kernel, see Figure \ref{fig:attn_visualization}.
Here, we use the slope to denote the strength of the second-order interaction $\mathcal{H}^{(g)}$ over first-order interaction $H^{(l)}$, indicating the impact of neighbor points towards linear correlation captured by the gating function $h^{(g)}$.
The learned second-order point interaction demonstrates a strong geometric harmony with Octahedron kernel dispositions (i.e., 1-4-2 groups among symmetrical axis):
the two vertices that are the most distant from the center point show the most significant interaction (i.e., $\mathcal{H}^{(g)}$) towards global neighborhood information $\mathcal{H}^{(l)}$. This supports more robust recognition results, as these kernel points are involved in addressing the boundary effects of sparse and fast-changing point clouds.
The 4 vertices connecting the square plane have the least significant interaction, suggesting a smoother learning on a denser point distributions to capture the features of point majorities.
The center kernel point admits most information from centralized points, and thus co-adapts to global neighbor information and yields a moderate interaction strength.

\subsection{Effectiveness of DS Predictor}
We show that even marginal Kendall $\tau$ improvement (i.e., 0.008) in DS predictor leads to significant quality improvement of NAS-crafted models.
We establish a random search baseline by selecting the top-5 models from the outcome of 1K random samples, and establish predictor-based NAS baseline by selecting top-5 models using the aforementioned runner-up predictor: Sparse predictor in Table \ref{tab:predictor_ablation}.
All models are evaluated on classification and segmentation benchmarks using the same 100-epoch training pipeline. 
We report the mean and standard deviation of performance on SemanticKITTI in Table \ref{tab:predictor_ablation}.

\begin{table}[t]
    \vspace{0.5em}
    \caption{Comparison of various predictor-based NAS methods with Random Search baseline. Results are derived from the top-5 NAS models discovered by each method.}
    \vspace{-0.5em}
    \begin{center}
    \scalebox{0.95}{
    \begin{tabular}{|c|cccc|cccc|}
    \hline
     \multirow{2}{*}{\textbf{Search Method}} & \multicolumn{4}{|c|}{\textbf{SemanticKITTI mIOU (\%)}}\\
      & \textbf{mean} & \textbf{std} & \textbf{max} & \textbf{min}\\
    \hline
    \hline
    Random & 53.54 & 1.19 & 54.68 & 51.96 \\
    Sparse predictor &  54.90 & 0.49 & 55.28 & 54.2 \\
    Dense-Sparse predictor & \textbf{56.10} & 0.85 & 56.83 & 54.87 \\
    \hline
    \end{tabular}}
    \label{tab:predictor_ablation}
    \end{center}
    \vspace{-2.5em}
\end{table}

Results demonstrate that: (1) without introducing extra architecture samples, predictor-based NAS can outperform random search by up to 1.4\% mIOU on SemanticKITTI. (2) Even with 0.07$\times$10$^{-2}$ lower MSE, the NAS-crafted models via DS predictor can outperform NAS-crafted models via Sparse predictor by a large margin (i.e., up to 1.2\% mIOU on SemanticKITTI). 
This shows the significance of improving the design of predictor and enabling superior neural architecture representations for predictor-based NAS.
\section{Conclusion}
In this work, we present a new paradigm, Joint \textbf{P}oint \textbf{I}nteraction-\textbf{D}imension \textbf{S}earch for 3D Point Cloud (PIDS), to jointly search point interactions and point dimensions on two axes of 3D point operators.
PIDS innovates high-order point interactions and efficient point dimensions to suit geometry and density heterogeneity of 3D point cloud data, and establishes a search space to carry joint exploration.
PIDS utilizes predictor-based NAS and proposes a novel Dense-Sparse predictor to enhance the quality of predictions and the ranking of candidate networks.
Dense-Sparse predictor utilizes enhanced priors to encode heterogeneous search components, and interact discrete/continuous architecture representations through a cross-term dot product.
Results on 2 semantic segmentation benchmarks justify the state-of-the-art performance of NAS-crafted PIDS models, and its good transferability of searched models.

\noindent \textbf{Acknowledgement.}
This project is supported in part by NSF-2112562, NSF-1937435, NSF-2148253, ARO W911NF-19-2-0107, and CAREER-2048044.

{\small
\bibliographystyle{ieee_fullname}
\bibliography{egbib}
}

\newpage
\appendix

\end{document}


\title{
\textit{Supplementary Material for}

PIDS: Joint Point Interaction-Dimension Search for 3D Point Cloud

}

\maketitle
\thispagestyle{empty}

\setcounter{section}{9}
\setcounter{table}{8}
\setcounter{figure}{4}

In this appendix, we first evaluate PIDS on ModelNet40 to study its effect on classification benchmarks.
Then, we elaborate the detailed structure of hand-crafted PIDS, built upon efficient 3D point operators within the PIDS search space.
Finally, we discuss the detailed configuration of Dense-Sparse predictor. Our code is publicly available \href{https://github.com/lordzth666/WACV23_PIDS-Joint-Point-Interaction-Dimension-Search-for-3D-Point-Cloud}{here}.

\subsection{Evaluation on ModelNet40 Classification}
We run PIDS on ModelNet40 classification benchmark to search for highly representative classification models.
On ModelNet40, a single training batch in ModelNet40 contains 16 sub-sampled point clouds.
We adopt a similar training pipeline following the original KPConv~\cite{thomas2019kpconv} paper yet incorporates SimpleView~\cite{goyal2021revisiting} RSCNN training protocol for fair comparison.
Specifically, the SimpleView-RSCNN protocol adopts random scaling \& translation for data augmentation, employ cross-entropy loss to optimize, and utilize a voting scheme on the best test model to best exploit the potential of searched model.
\begin{table}[h]
    \begin{center}
    \caption{Overall Accuracy (OA) on ModelNet40. $^*$: Use native protocol instead of SimpleView RSCNN evaluation protocol.}
    \scalebox{1.0}{
    \begin{tabular}{|c|c|c|}
        \hline
         \multirow{2}{*}{\textbf{Architecture}} &  \textbf{Params} &  \textbf{OA} \\
         &  \textbf{(M)} & \textbf{(\%)}  \\
         \hline
         PointCNN~\cite{li2018pointcnn} & 0.60 & 92.2 \\
         PointConv~\cite{wu2019pointconv} & - & 92.5 \\
         SPH3D-GCN$^{*}$~\cite{lei2020spherical} & 0.8  & 92.1 \\
         FPConv$^{*}$~\cite{lin2020fpconv} & 2.1 & 92.5 \\
         RSCNN~\cite{liu2019relation} & 1.3 & 92.5 \\
         DGCNN~\cite{wang2019dynamic} & 1.8 & 92.8 \\
         KPConv~\cite{thomas2019kpconv} & 14.9 & 92.9 \\
         SimpleView~\cite{goyal2021revisiting} & 0.80 & 93.2 \\
         PointNet++~\cite{qi2017pointnet++} & 1.48 & 93.3 \\
         \hline
         PIDS (second-order) & 1.25 & 92.6 \\
         \hline
         PIDS (NAS) & \textbf{0.56} & 93.1 \\
         PIDS (NAS, 2$\times$) & 1.21 & \textbf{93.4} \\
         \hline
    \end{tabular}
    }    
    \label{tab:result_modelnet40}
    \end{center}
    \vspace{-2em}
\end{table}

We demonstrate the evaluation results on the testing dataset of ModelNet40 in Table \ref{tab:result_modelnet40}.
PIDS explores more efficient model designs with higher performance, out-performing KPConv and other prior arts~\cite{goyal2021revisiting} with up to 1.3\% OA while being up to 12$\times$ smaller in size.

\subsection{Structure of First-order PIDS}
\begin{table}[h]
\caption{Structure of hand-crafted PIDS (first-order). "1/2" strides means up-sampling by 2$\times$, and "O" denotes "Octahedron" kernel with 7 kernel points. All point operators employ a first-order point interaction.}
\begin{center}
\scalebox{0.95}{
    \begin{tabular}{|c|c|c|c|c|c|c|c|}
        \hline
        \multirow{2}{*}{\textbf{Point-Op}} & \multirow{2}{*}{\textbf{Depth}} & \multirow{2}{*}{\textbf{Stride}} & \textbf{In} & \textbf{Out} & \multirow{2}{*}{\textbf{E}} & \multirow{2}{*}{\textbf{K}} \\
        & & & \textbf{Width} & \textbf{Width} & & \\
        \hline
         \textbf{1} & 1 & 1 & 16 & 16 & 1 & O \\
         \textbf{2} & 2 & 2 & 16 & 24 & 3 & O \\
         \textbf{3} & 3 & 2 & 24 & 32 & 3 & O \\
         \textbf{4} & 4 & 2 & 32 & 64 & 3 & O \\
         \textbf{5} & 3 & 1 & 64 & 96 & 3 & O \\
         \textbf{6} & 3 & 2 & 96 & 160 & 3 & O \\
         \textbf{7} & 1 & 1 & 160 & 320 & 3 & O \\
         \textbf{8} & 1 & 1/2 & 416 & 160 & 3 & O \\
         \textbf{9} & 1 & 1/2 & 160 & 96 & 3 & O \\
         \textbf{10} & 1 & 1/2 & 96 & 64 & 3 & O \\
         \textbf{11} & 1 & 1/2 & 64 & 32 & 3 & O \\
         \hline
    \end{tabular}
    }
\end{center}
\label{tab:pids_1st_order}
\vspace{-2em}
\end{table}
We follow the layer organization to manually craft first-order PIDS (Table 2), and demonstrate the detailed architecture in Table \ref{tab:pids_1st_order}, echoing the design of search components in Section 3.3.
The hand-crafted architecture contains a total of 11 point operators (Point-Op). 
Among them, 7 point operators serve as a classification backbone, and 4 additional point operators serve as a semantic segmentation head. 
The positional setting (i.e., depth, block width) of hand-crafted PIDS strictly follows MobileNet-V2, which is considered as a good hand-crafted model design with remarkable efficiency.
For structural settings, we employ an expansion factor (E) of 3 for all point operators except for the first one.
We leverage the 7-point Octahedron layout for each point operator for design efficiency. 
As a result, first-order PIDS serves as a strong baseline to compare with, when we evaluate the performance of NAS-crafted PIDS models.

\subsection{Details of PIDS Search Space}
We present the detailed settings of the PIDS interaction-dimension search space in Table \ref{tab:search_cfg}. 
We jointly search the point interaction (kernel size and interaction type) and point dimension (block depth, block width, expansion factor) to find the best architecture.

\begin{table*}[t]
\begin{center}
    \caption{Configuration of the PIDS Search Space. "1/2" strides means up-sampling by 2$\times$.}
    \scalebox{0.825}{
    \begin{tabular}{|c|c|c|c|c|c|c|c|}
    \hline
        \textbf{Hierarchy} & \textbf{Stage}  & \textbf{Strides} & \textbf{Order Type}  & \textbf{Kernel Type} & \textbf{Depth} & \textbf{Expansion Factor} & \textbf{Width} \\
        \hline
         & \textbf{1} & 1 & first-order & Tetrahedron, Octahedron, Icosahedron & 1  & 1 & 16 \\
         & \textbf{2} & 2 & first-order & Tetrahedron, Octahedron, Icosahedron & 2, 3 & 2, 3, 4 & 16, 24 \\
         & \textbf{3} & 2 & first-order/second-order  & Tetrahedron, Octahedron, Icosahedron & 2, 3, 4 & 2, 3, 4 & 24, 32 \\
        \textbf{Backbone} & \textbf{4} & 2 & first-order/second-order &  Tetrahedron, Octahedron, Icosahedron & 3, 4, 5 & 2, 3, 4 & 24, 32, 40 \\
         & \textbf{5} & 1 & first-order/second-order & Tetrahedron, Octahedron, Icosahedron  & 2, 3, 4 & 2, 3, 4 & 40, 56, 72 \\
         & \textbf{6} & 2 & first-order/second-order  & Tetrahedron, Octahedron, Icosahedron & 3, 4, 5 & 2, 3, 4 & 64, 80, 96 \\
         & \textbf{7} & 1 & first-order/second-order  & Tetrahedron, Octahedron, Icosahedron & 1 & 2, 3, 4 & 160 \\
         \hline
         & \textbf{8} & 1/2 & first-order/second-order &  Tetrahedron, Octahedron, Icosahedron & 1  &  2, 3 & 64, 80, 96 \\
         \textbf{Segmentation} & \textbf{9} & 1/2 & first-order/second-order & Tetrahedron, Octahedron, Icosahedron & 1  & 2, 3 & 40, 56, 72 \\
         \textbf{Head} & \textbf{10} & 1/2 & first-order/second-order  & Tetrahedron, Octahedron, Icosahedron & 1  & 2, 3 & 24, 32, 40 \\
         & \textbf{11} & 1/2 & first-order & Tetrahedron, Octahedron, Icosahedron & 1 & 2, 3 & 16, 24 \\
         \hline
    \end{tabular}
    }    
    \label{tab:search_cfg}
    \end{center}
    \vspace{-2em}
\end{table*}

\subsection{Design of Dense-Sparse Predictor}
We manually configure the architectural hyperparameters of our Dense-Sparse predictor without delicate architecture engineering.
Specifically, we employ a 3-layer MLP with 64-128-256 layers as the dense architecture to extract dense neural architecture representations for both positional organizations and structural settings.
The overall architecture that is responsible for processing fused features after dense-sparse interaction is a 2-layer MLP with 256-256 layers. We utilize ReLU as the activation function and apply Dropout of 0.5 before the final regression head to mitigate overfitting. To ensure fair comparison in Table 3, both the Dense predictor and the Sparse predictor also adopt a 2-layer MLP with 256-128 units to keep the same maximum projection dimension as the Dense-Sparse predictor.

{\small
\bibliographystyle{ieee_fullname}
\bibliography{egbib}
}

\newpage
\appendix